\documentclass[letterpaper, 10 pt, conference]{ieeeconf}  

\IEEEoverridecommandlockouts                              

\usepackage{graphics} 
\usepackage{epsfig} 
\usepackage{mathptmx} 
\usepackage{times} 
\usepackage{amsmath} 
\usepackage{amssymb}  
\usepackage{siunitx}
\usepackage{subcaption}
\DeclareSIUnit\px{px}
\usepackage{xcolor}
\usepackage{hyperref}

\title{\LARGE \bf Key Point-based Orientation Estimation of Strawberries for Robotic Fruit Picking}

\author{Justin Le Louëdec$^{1}$ and Grzegorz Cielniak$^{1}$ 
\thanks{$^{1}$Lincoln Centre for Autonomous Systems, University of Lincoln, Brayford Way, Brayford Pool, Lincoln LN6 7TS, United Kingdom
        {\tt\small \{jlelouedec,gcielniak\}@lincoln.ac.uk}}%
}

\begin{document}

\maketitle
\thispagestyle{empty}
\pagestyle{empty}

\begin{abstract}
Selective robotic harvesting is a promising technological solution to address labour shortages which are affecting modern agriculture in many parts of the world. For an accurate and efficient picking process, a robotic harvester requires the precise location and orientation of the fruit to effectively plan the trajectory of the end effector. The current methods for estimating fruit orientation employ either complete 3D information which typically requires registration from multiple views or rely on fully-supervised learning techniques, which require difficult-to-obtain manual annotation of the reference orientation.

In this paper, we introduce a novel key-point-based fruit orientation estimation method allowing for the prediction of 3D orientation from 2D images directly. The proposed technique can work without full 3D orientation annotations but can also exploit such information for improved accuracy. We evaluate our work on two separate datasets of strawberry images obtained from real-world data collection scenarios. Our proposed method achieves state-of-the-art performance with an average error as low as $8^{\circ}$, improving predictions by $\sim30\%$ compared to previous work presented in~\cite{wagner2021efficient}. Furthermore, our method is suited for real-time robotic applications with fast inference times of $\sim30$ms.

\end{abstract}

\begin{figure}[!ht]
    \begin{center}
        \includegraphics[width=\linewidth]{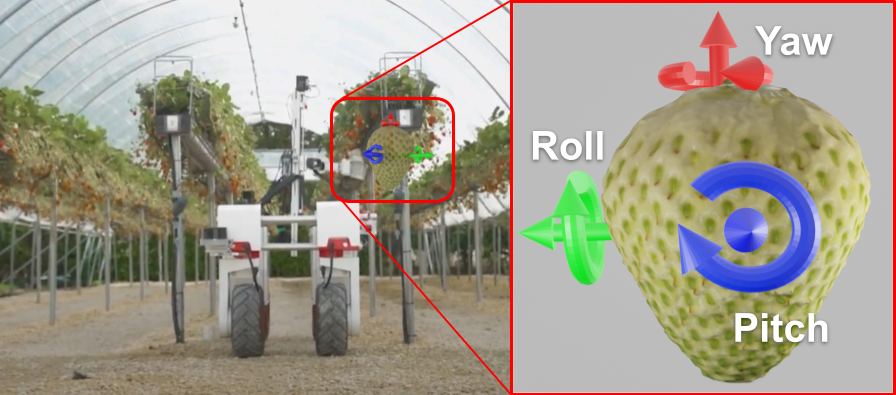} 
    \end{center}
    \caption{An example strawberry harvesting robot with an example strawberry indicating the Roll, Pitch, Yaw (RPY) orientation representation assumed in this work.}
    \label{fig:anglesexamplesvisu}
\end{figure}

Automation through robotisation of the agricultural sector is seen as a promising solution to the socio-economic challenges faced by this industry. A key application that would benefit most from automation, which still relies on manual human labour, is selective crop harvesting. Cultivated strawberries (\textit{Fragaria x ananassa}) are a perfect example of a crop with a recent significant increase in demand but affected by labour shortages. Like many varieties of fruit, they have several characteristics, rendering their harvesting challenging for robots. To perform precise manipulation and grasping of the harvestable crop, robotic systems require precise information about the location and pose of the crop (see Fig.~\ref{fig:anglesexamplesvisu}). Whilst detecting and localising strawberry crop has been well-studied in prior work (e.g., ~\cite{kirk2020b,ge2019instance}), inferring their precise pose is still an ongoing challenge. Typical approaches for estimating the location and pose of the fruit rely on a combination of 2D images together with 3D information (e.g.,~\cite{xiong2021improved}) and require additional projection steps and direct operations on point clouds which add significant computation overhead. 

In this paper, we present a method that enables precise fruit pose estimation directly from 2D images allowing for bypassing computationally expensive transformation and estimation steps. A similar approach has been originally proposed by \cite{wagner2021efficient} which estimates the fruit pose by regressing the orientation direction vector directly from an image in a fully supervised manner using a learnt feature extractor and regression layer. The method is characterised by fast inference times but also reduced accuracy caused by occasional failures which are difficult to attribute to any particular part of the architecture. In addition, the supervised approach relies on difficult-to-obtain high-quality orientation annotations making the method impractical for wider use. In contrast, the proposed method employs a two-stage approach first predicting the localisation of two characteristic key points of the strawberry fruit which are then used for pose estimation. This two-stage approach, inspired by prior work in human pose estimation (i.e.,~\cite{newell2016stacked}), leads to improved accuracy and a simplified training regime thanks to the straightforward annotation of key points. In particular, the contributions of this paper include:
\begin{itemize}
    \item a new method for efficient estimation of fruit orientation directly from colour images based on key-point detectors;
    \item an improved roll angle estimation method based on learnt image features and the estimated key point information;
    \item evaluation of the proposed fruit pose estimation system on two different datasets of strawberry images consisting of high-resolution reconstructed models of fruit and data collected directly in the field demonstrating the improvements to the state of the art.
\end{itemize}

\section{Related work}

Obtaining accurate pose estimation of objects, and especially the orientation component is essential for enabling precise manipulation and grasping tasks such as robotic picking~~\cite{harrell1990robotic,baeten2008autonomous,liu2022robotic}. Whilst early approaches utilised handcrafted image descriptors~\cite{munoz2016fast,park2017mutual} for that task, most recent research focuses on learning approaches exploiting the existing extensive datasets. The examples include work making use of multiple-views for precise orientation and category estimation such as in~\cite{kanezaki2018rotationnet}, through iterative refinement of object's pose using Gated Recurrent Unit (GRU) operators as in~\cite{lipson2022coupled} or by employing transformer architectures as in~\cite{jantos2022poet}. The majority of the state of that art methods, however, rely on precise and difficult-to-obtain annotations and therefore there is a big interest in methods which can reduce that requirement for example through the use of simulation for training as in Sim2Real~\cite{zhong2022sim2real}. When accurate 3D point clouds are available, the use of 3D geometric insights can be used to improve orientation prediction across the whole categories of objects~\cite{di2022gpv}. All these techniques, however, do not focus on a specific application and rely on benchmark datasets of non-organic objects exclusively (e.g., YCB-V dataset~\cite{xiang2017posecnn}), making their usefulness for agri-robotics applications more difficult to assess.

Object pose estimation for robotic manipulation and grasping has been researched thoroughly over the years. For example, the approach in~\cite{deng2020self} is proposing the use of simulation and self-supervised training through manipulating objects to identify their pose. There are also methods relying on point cloud data and improved descriptors which were applied to obtain better accuracy of object orientation for robotic bin picking~\cite{cui20226d}. Crop pose estimation for robotic harvesting has also gained recent interest with various applications in different crops. For example, the approach presented in~\cite{lin2019guava} estimates the pose of guava fruit from point cloud data obtained from an RGBD sensor by segmenting plant components (i.e. fruit and branches) and combining their relative pose. Other methods, such as~\cite{guo2020pose}, propose the refinement of the fruit pose by registering a 3D reconstruction of the captured point cloud to offline templates of the identified fruit ~\cite{guo2020pose}. The method presented in~\cite{kim2022tomato} predicts the crop detection bounding boxes, maturity, pose and precise stem orientation to identify the optimal cutting point for tomatoes, obtaining more accurate and thorough information for harvesting.

For strawberries, recent work introduced a learning-based regression of the orientation vectors of the fruits from a single colour and, optionally, depth image~\cite{wagner2021efficient}. While achieving state-of-the-art results, the method's accuracy is affected by occasional failures which due to the dimensional difference between the images and the output 3D orientation vector are difficult to analyse. In addition, the fully-supervised nature of the method requires accurate annotation of the fruit orientation in images which are complex to obtain without additional geometrical information about the fruit size and shape.

In contrast to the state of the art, our method employs key-point detectors for estimating 3D fruit orientation directly from 2D images. The proposed technique can work without full 3D orientation annotations but can also exploit such information for improved accuracy.

\section{The Approach}
The proposed method consists of a set of key components which include a learnt key point detector, orientation calculation and an optional, learnt component for improved estimation of the orientation. In this work, we assume the ``roll-pitch-yaw'' representation as per~\cite{wagner2021efficient}. In the proposed application (see Fig.~\ref{fig:anglesexamplesvisu}), the yaw angle is irrelevant due to the symmetry and non-uniformity of the strawberries. Thus, we simplify the definition of the orientation to two angles: pitch ($\phi \in \left[-180,180\right]$) and roll ($\theta \in \left[0,90\right]$). Both angles can be derived numerically from the detected crop key points in 2D, but in addition, we demonstrate that the calculation of $\theta$ can be regressed from the image and the key points, leading to improved results when compared to the direct numerical formula. The main advantage of key points is the ease of their annotation in images which involves the marking of a single-pixel location only.

In our case, the two ``top'' and ``tip'' key points represent projections of the stem attachment point and the extreme point of the fruit, respectively onto the image plane. Due to the fruit growing conditions and typical harvesting robot camera configuration, the tip is always located in space between being parallel to the image plane or pointing towards the camera. At the same time, the top might become obstructed by the crop and invisible in the image. In such a case, its projection on the image plane relative to the centre of mass is a good indication of the fruit's size. An example demonstrating the appearance of key point location under different fruit rotations is presented in Fig.~\ref{fig:distancestheta}. 
    
\subsection{Key point detection}

\begin{figure}[!ht]
    \begin{center}
        \includegraphics[width=\linewidth]{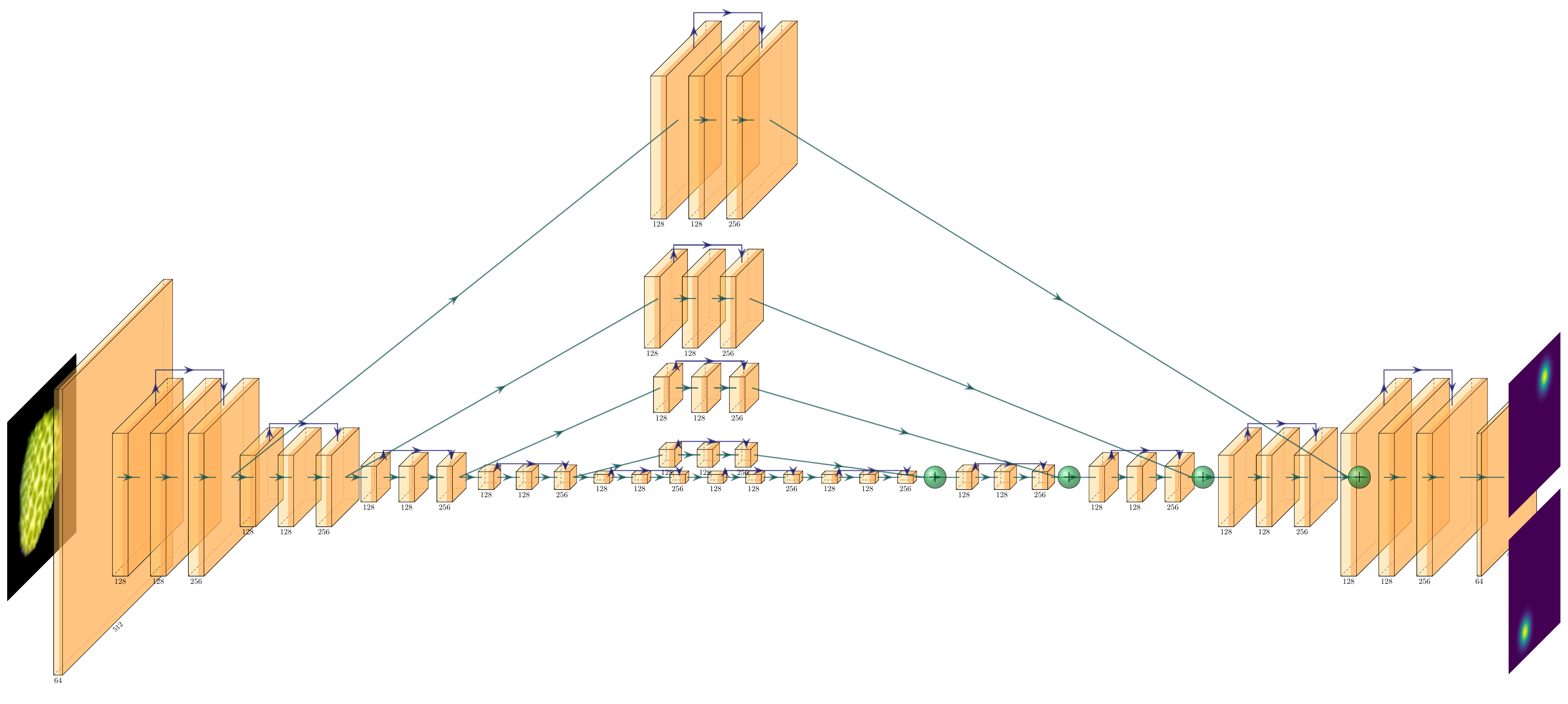} 
    \end{center}
    \caption{An hourglass network composed of a succession of residual modules consisting of convolution layers with skip connection from input to output as in~\cite{he2015deep}. The input is the colour RGB image, and the output is two heat maps indicating the location of each key point.}
    \label{fig:hourglass}
\end{figure}

\begin{figure}[!ht]
    \begin{center}
        \includegraphics[width=\linewidth]{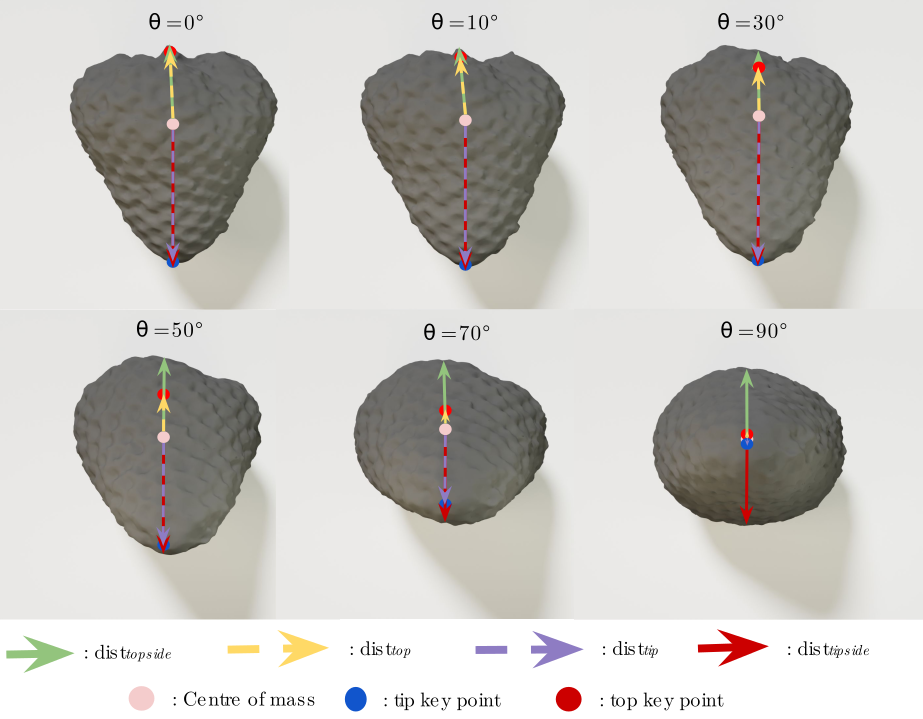} 
    \end{center}
    \caption{The location of key points and relative distances used for orientation calculation under changing values of $\theta$.}
    \label{fig:distancestheta}
\end{figure}

We propose using a method inspired by human pose estimation literature for predicting the position of the key points. For each strawberry image, we estimate two heat maps corresponding to the ``top'' and ``tip'' key points. The top and tip key points correspond to the relative position of the stem attachment point and outer part of the berry respectively projected onto the image plane (see Fig.~\ref{fig:distancestheta}). We use the method presented in~\cite{newell2016stacked} composed of a stack of $S$ small hourglass networks (see Fig.~\ref{fig:hourglass}). For our scenario, we choose experimentally $S=8$ with one final convolution layer followed by a sigmoid function for each hourglass network. This last activation function forces the network outputs within the range $[0,1]$ and can be interpreted as the likelihood of the location of the key points. The output of an intermediary hourglass module is combined with its feature map and input for the next module in the stack. 	
  
During training, we use the average binary cross-entropy as the loss function $L_{K}=-ylog(f(x))-(1-y)log(1-f(x))$, where$f(x)$ is the output of each stacked hourglass and $y$ the ground truth heat map for the key point. During inference, the key point location corresponds to the image coordinates at the maximum value selected from the $S=8$ predicted heat maps for each key point. The ground truth heatmaps consist of 2D Gaussian kernels ($\sigma=2.0$) applied to images containing each annotated key point location.

\subsection{Key point-based orientation}
We define the coordinates of the key points as $A$ and $B$ which correspond to the top and tip of the strawberries respectively. The roll $\phi$ can be calculated in a straightforward manner: $\phi = \arctan(\frac{y_{BA}}{x_{BA}})$, where $\vec{BA} = (x_{BA},y_{BA})$ is a vector between the both key points. 

We propose to derive pitch $\theta$ from the relative position of the key points to the centre of mass and also to the geometric outline of the strawberry. As the fruit rotates, the two key points get closer/further to each other and it is therefore possible to correlate their relative distance to the orientation of the berry. To this end, we define the following four distances with respect to the centre of mass: 
\begin{itemize}
    \item $d_{top}$ and $d_{tip}$ representing the distance to the top and tip key points, respectively;
    \item $d_{topside}$ and $d_{tipside}$ is the distance to the fruit's contour following the straight line intersecting the top and tip key points, respectively. 
\end{itemize}
We then use these measures to derive normalised distances to the key points as ratios $\hat{d}_{top}=\frac{d_{top}}{d_{topside}}$ and $\hat{d}_{tip}=\frac{d_{tip}}{d_{tipside}}$. The key point locations and corresponding distances relative to the centre of mass for different values of $\theta$ are illustrated in Fig.~\ref{fig:distancestheta}.

Due to the specific shape and the skewed centre of mass of strawberries, the 3D rotation of the fruit results in an elliptical trajectory of the key points in the image space which we model using a simple square root model (see Eq.~\ref{eq:pseudoyaw}). With the changing pitch angle $\theta$, the $\hat{d}_{top}$ decreases fast until a specific value ($\sim\theta=50^{\circ}$ in our case), whilst $\hat{d}_{tip}$ is not affected as much (see Fig.~\ref{fig:distancestheta}). The distance $\hat{d}_{tip}$ is, however, a good indicator for $\theta$ values above that threshold. Finally, with the image resolution normalised, we can use the length of the berry $T$ as a threshold between the phase of importance for $\hat{d}_{top}$ and $\hat{d}_{tip}$. With these considerations, we define $\theta$ as: 
\begin{equation}\label{eq:pseudoyaw}
    \theta = 
    \begin{cases}
        \sqrt{\hat{d}_{top}}\alpha & \quad \text{if}\quad d_{tt}>T,\\
        \sqrt{\hat{d}_{tip}}\omega+\sigma  & \quad \text{otherwise,}
    \end{cases}
\end{equation}
where $d_{tt}$ is the distance between the top and tip key points. The parameters $\alpha$, $\omega$ and $\sigma$ define the non-linear relationship between the relative key point distances and $\theta$. The values of these parameters are correlated with a typical shape of the fruit and in our case, for strawberries, these were tuned experimentally to $T=170.0$, $\alpha=54.0$, $\omega = 50.0$ and $\sigma = 40.0$. 

\subsection{Improved estimation of the pitch angle}

\begin{figure}[!ht]
    \begin{center}
        \includegraphics[width=\linewidth]{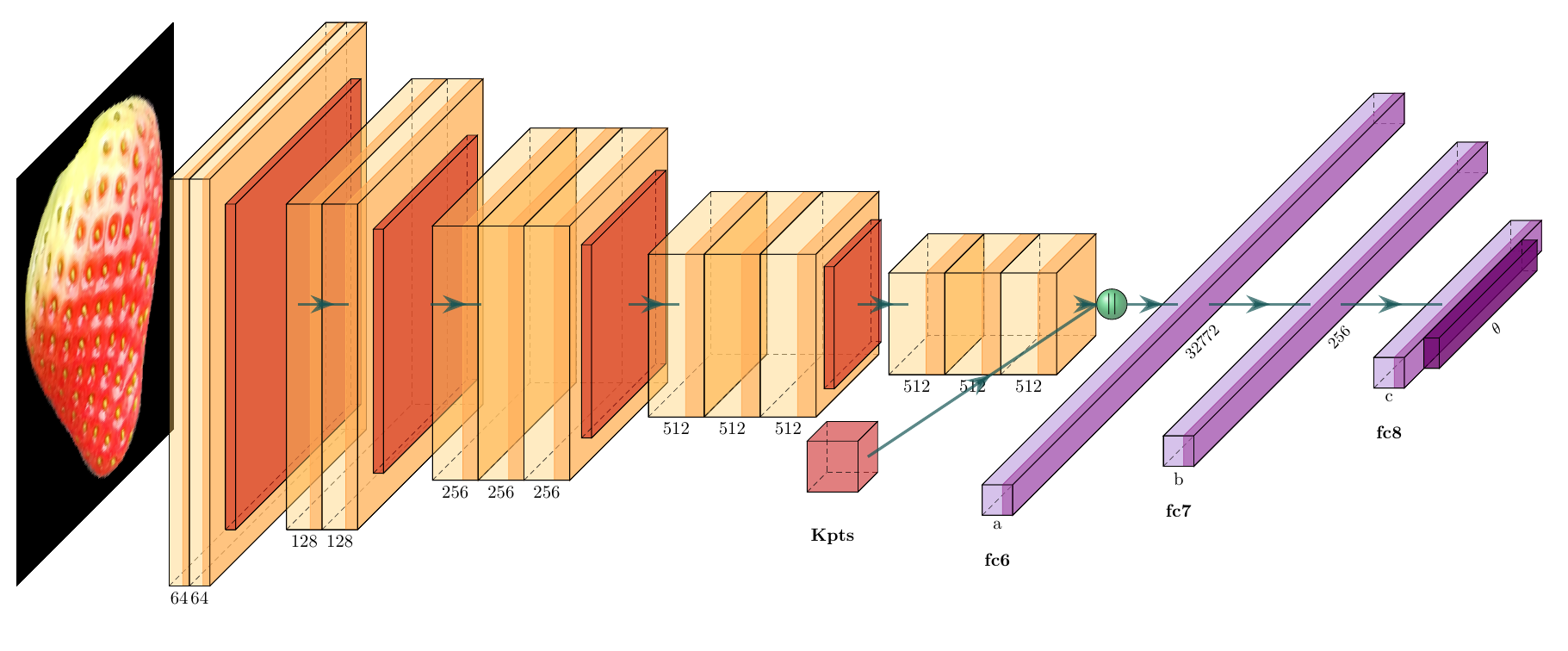} 
    \end{center}
\caption{A network architecture for predicting $\theta$ from the two key points and an input image. The feature extractor (orange) consists of convolutions (light shade) and max-pooling (dark shade) layers whilst the $\theta$ regressor (purple) includes fully-connected layers and the sigmoid activation function. The red cube represents the key point location input. The network outputs a single value $\theta \in \left[0,\frac{\pi}{2}\right]$.}
\label{fig:yawpredictor}
\vspace{-0.2cm}
\end{figure}

In general, obtaining the orientation ground truth for images is complex and not always possible. When this ground truth information is available (eg. with simulated data where object orientation is known), however, we propose a supervised method to predict the pitch angle $\theta$ from a strawberry image, which results in improved estimation results. Similarly to the previous work presented in~\cite{wagner2021efficient}, we use a pre-trained VGG16~\cite{simonyan2014very} architecture to extract relevant features from a strawberry image. However, the size of the regressor used in our architecture is larger to improve the predictive capabilities of the network. We then combine the extracted feature map with the detected key point locations before using a two-stage classifier to regress $\theta$. The architecture used to predict the $\theta$ is presented in Fig.~\ref{fig:yawpredictor}.
During training the loss is expressed as the mean squared error $L_{\theta}= (f(x)-y)^{2}$, where $f(x)$ is the output of the network and $y$ is the ground truth value for $\theta$.

\section{Evaluation Setup}

\subsection{Data collection}

To train and evaluate our method for orientation estimation of the strawberry fruit, we introduce two distinct datasets of strawberry images. $Straw_{2D}$ consists of strawberry images collected in-field and annotated with simple key point annotations whilst $Straw_{3D}$ includes images generated from high-quality 3D models of strawberries, which in addition to key point annotations, include also the full orientation ground truth.

\subsubsection{$Straw_{2D}$} 
\begin{figure}[!t]
\vspace{0.1cm}
\centering
 \begin{tabular}{c@{}c}
         \includegraphics[width=0.5\linewidth]{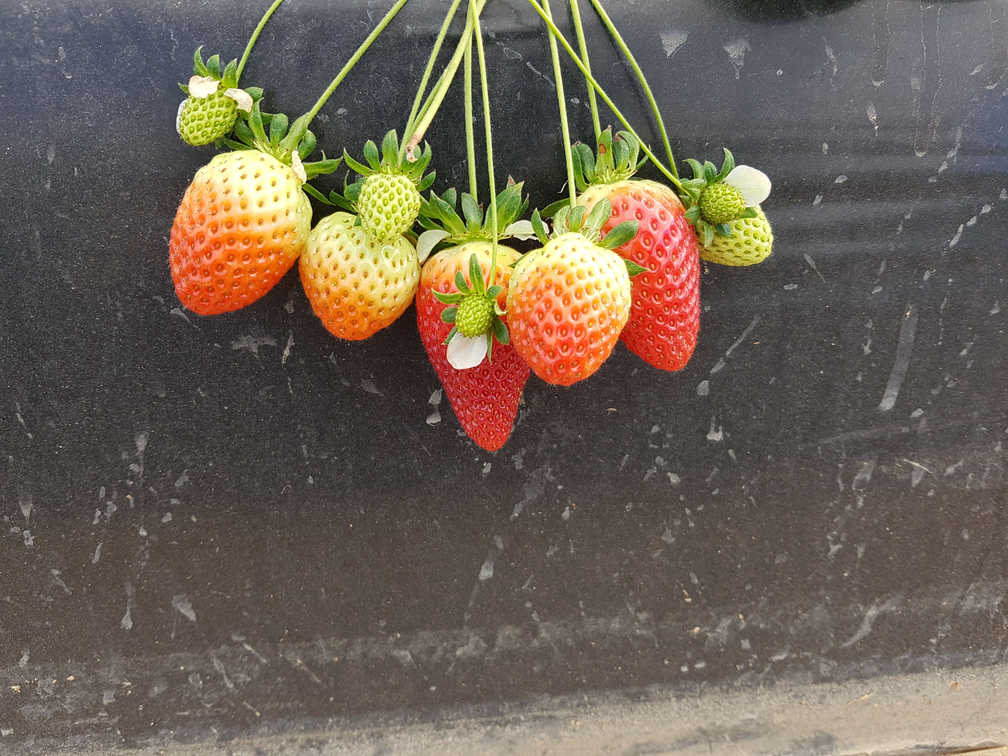}  & \includegraphics[width=0.5\linewidth]{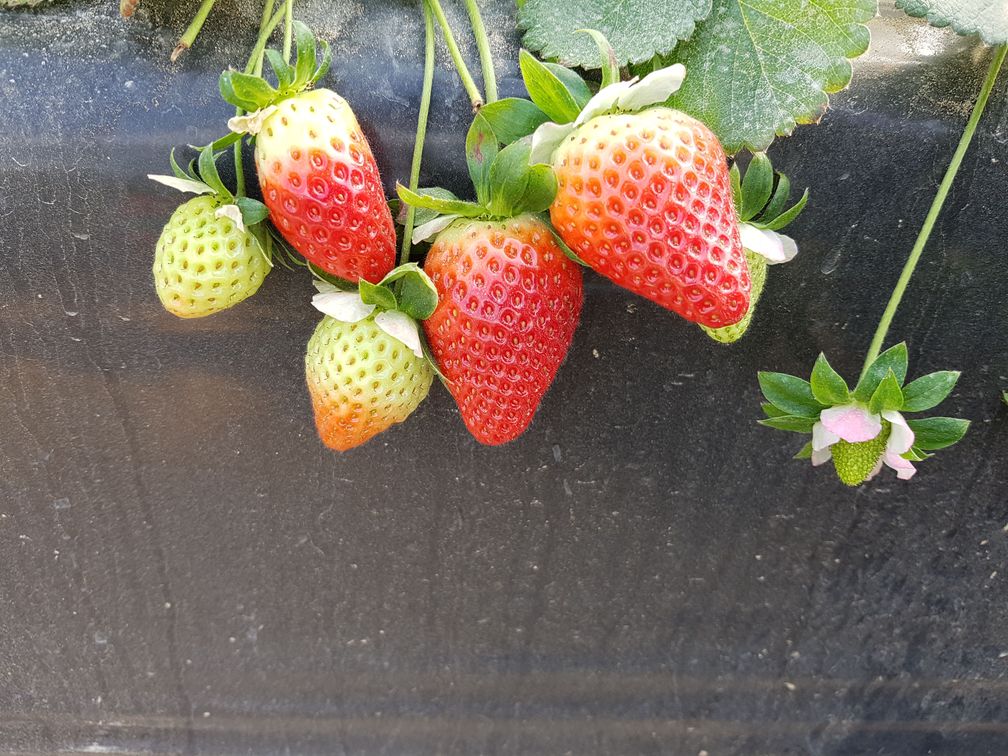} \\
         \multicolumn{2}{c}{\includegraphics[width=\linewidth]{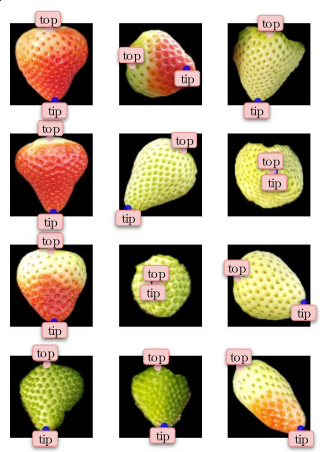}} \\
 \end{tabular}
\caption{Example original images from~~\cite{PEREZBORRERO2020105736} (top row) and the annotated instances of individual strawberries from $Straw_{2D}$ together with the annotation of the top (pink) and tip (blue) key points (bottom row).}\label{fig:set1example}
\vspace{-0.5cm}
\end{figure}

is based on the dataset provided in~\cite{PEREZBORRERO2020105736}, which includes high-resolution images of strawberries from 20 different plantations in Spain (see Fig.~\ref{fig:set1example}). From these images, 500 individual strawberries with minimal occlusion and of different shape, orientation and maturity stage were extracted and preprocessed so that background and calyx were masked out. The cropped square area around each berry was resized to the resolution of $\SI{256}{\px} \times \SI{256}{\px}$. The top and tip key points were then manually annotated by marking their projected location on the cropped image plane.

\subsubsection{$Straw_{3D}$}
is a dataset of strawberry images obtained from high-resolution 3D models. In addition to annotated key point locations, full orientation annotation is also available since the models can be rendered at the arbitrary pose. For creating these models, we have used multi-view stereo-photogrammetry which allows for reconstructing 3D point clouds from a set of images with unknown poses. The object is photographed from multiple views covering all sides of the object's geometry. For each image, features are computed using standard algorithms such as SIFT~\cite{lowe1999object} and are used to infer the relative pose between all the views. Pair-wise stereo depth prediction is then applied to produce a 3D object from every viewpoint. Photogrammetry creates precise 3D shapes and details without the need for distance measurements (e.g. typically provided by expensive lidar sensors) but requires significant computational resources which significantly increase with the number of images and the required precision. To create 3D shape ground truth of strawberries using photogrammetry, we use the Agisoft Metashape software~\cite{agisoft}.

\begin{figure}[!ht]
\centering
    \begin{subfigure}{\linewidth}
        \includegraphics[width=\hsize]{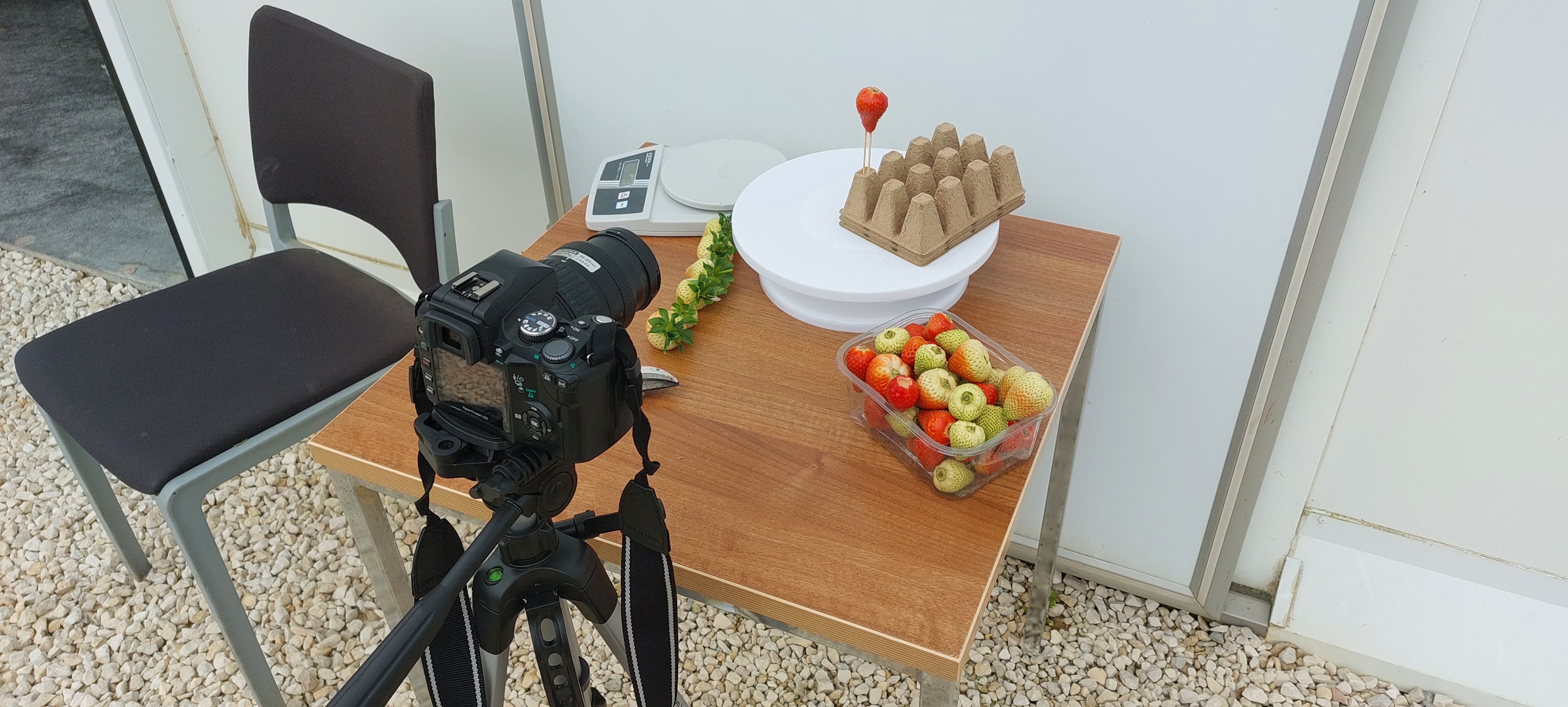}
        \caption{The photogrammetry setup used for capturing the 3D models of strawberries.}
        \label{fig:setups}
    \end{subfigure}
\vfill
\null
\vfill
    \begin{subfigure}{\linewidth}
        \begin{tabular}{c@{}c@{}c@{}c}
            \includegraphics[width=0.24\linewidth,height=0.45\linewidth]{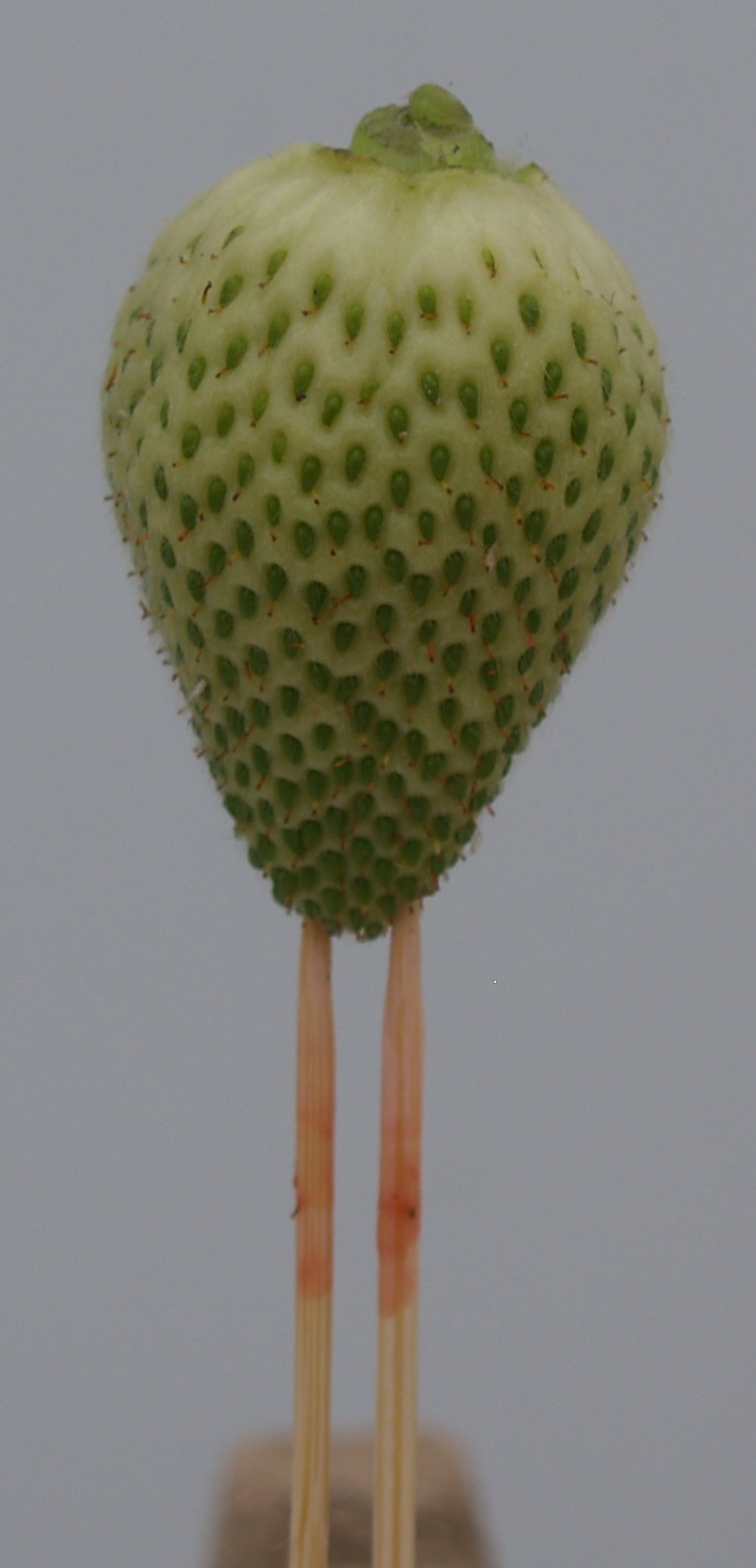}&
            \includegraphics[width=0.24\linewidth,height=0.45\linewidth]{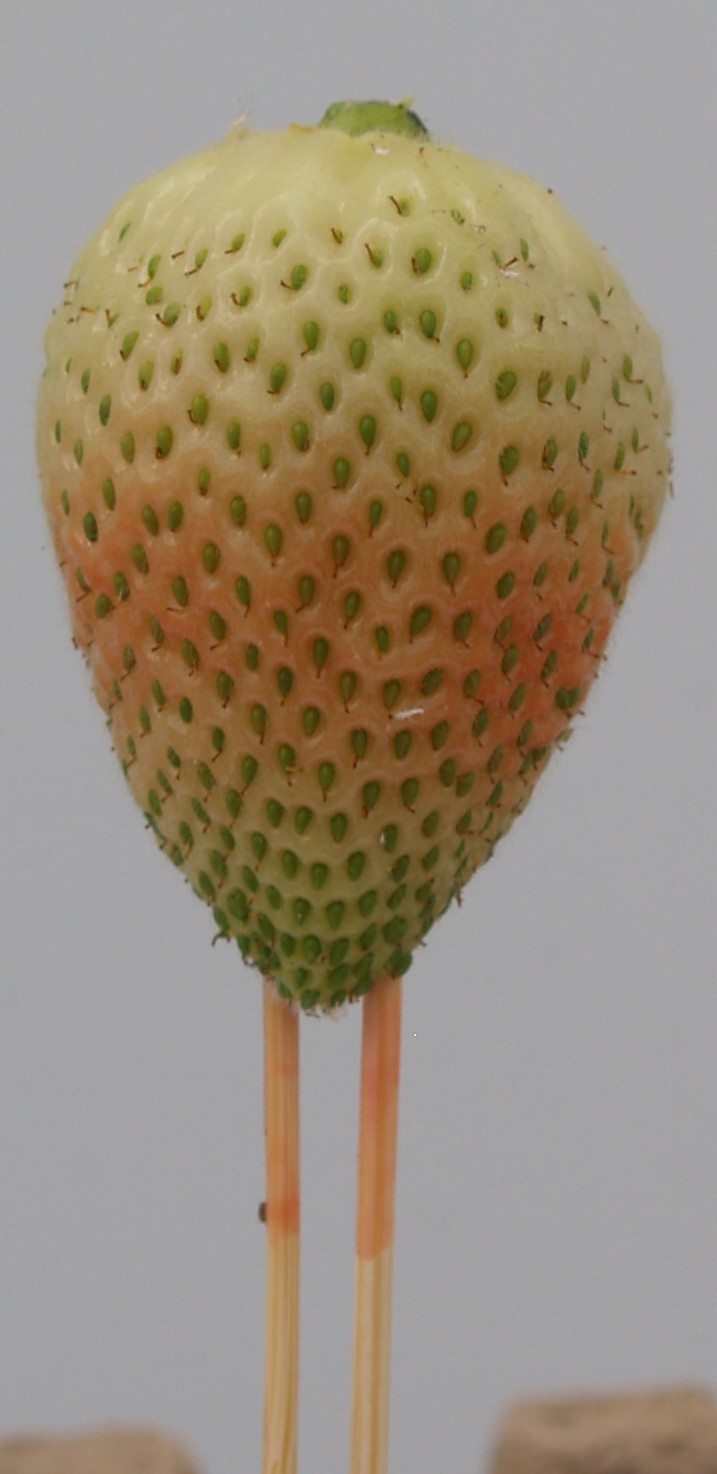}&
            \includegraphics[width=0.24\linewidth,height=0.45\linewidth]{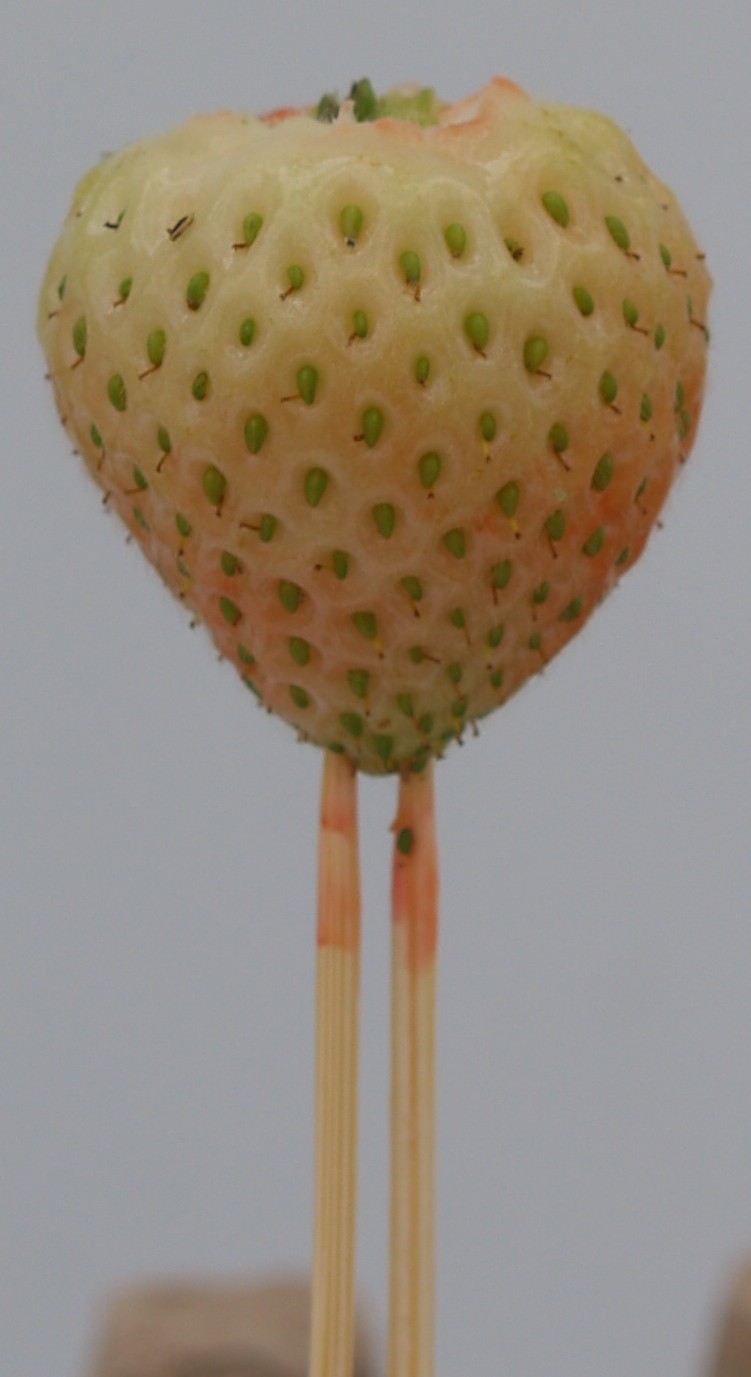}&
            \includegraphics[width=0.24\linewidth,height=0.45\linewidth]{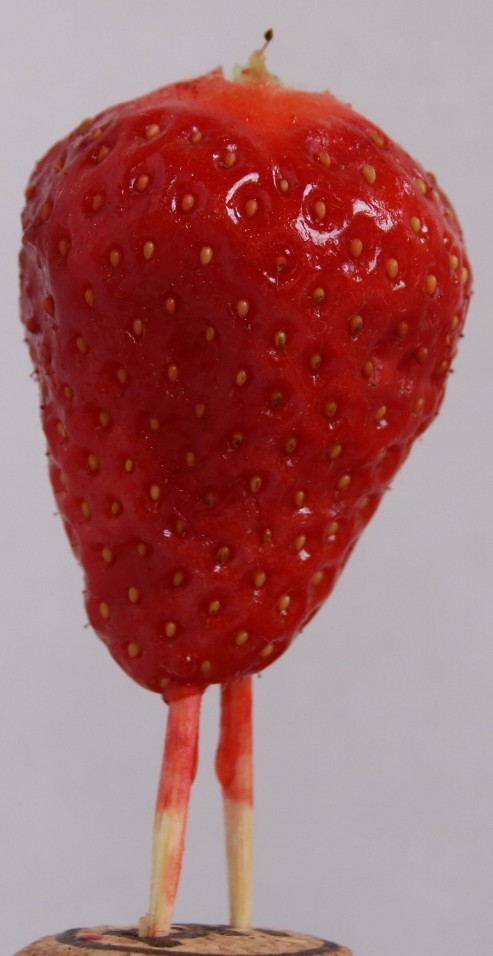}\\
        \end{tabular}
    \end{subfigure}
     \begin{subfigure}{\linewidth}
        \includegraphics[trim=0cm 6cm 0cm 0cm,clip,width=\hsize]{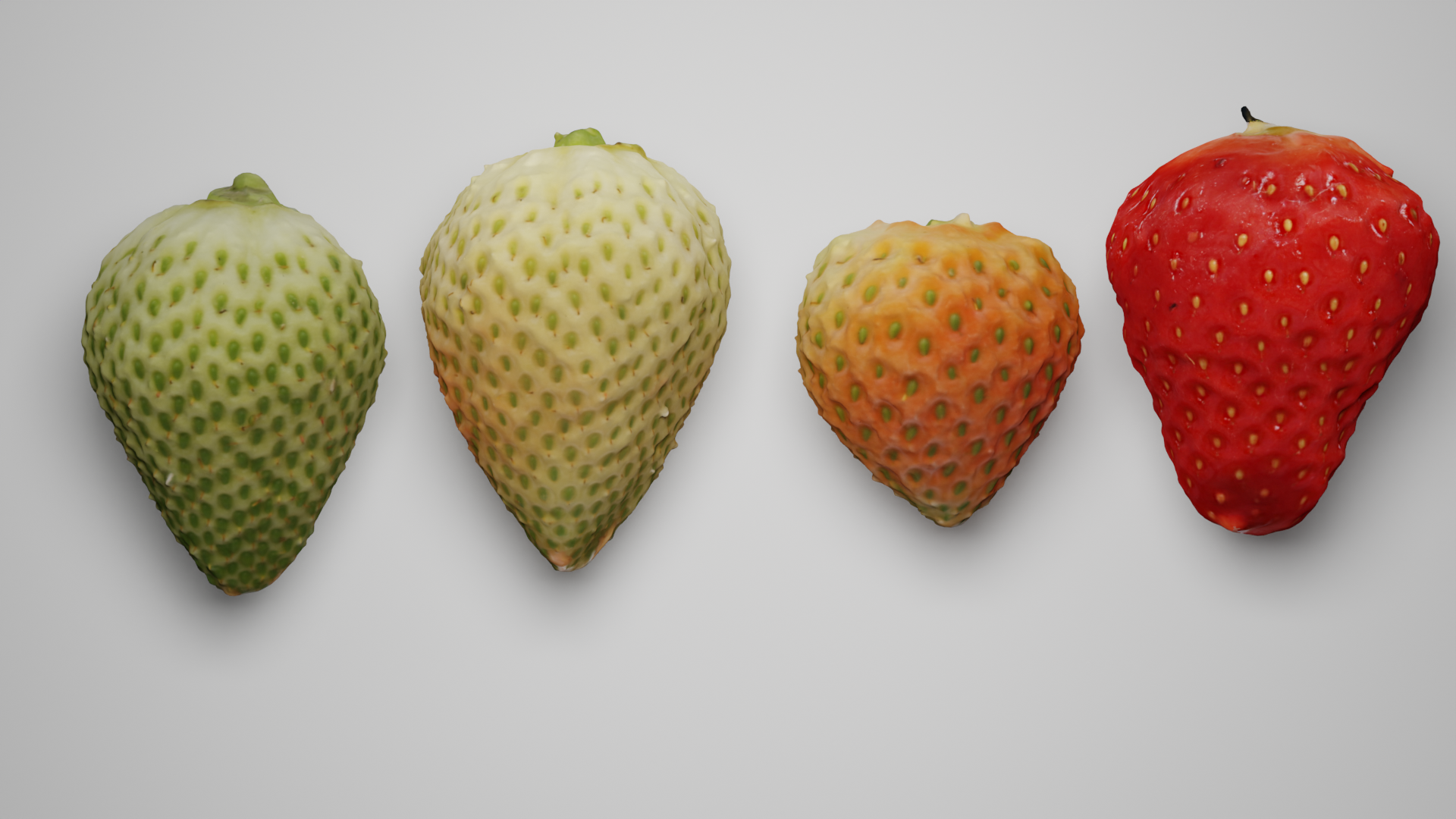}
    \end{subfigure}
\caption{Comparison between the images of real berries (top row) and their realistic 3D renders (bottom row).}\label{fig:ours_examples}
\end{figure}

Our setup for capturing the 3D models consists of a high-resolution camera (``Olympus E-50'' with a focal length of 45 mm) mounted on a tripod and a manually rotated table carrying a strawberry fixed with two picks preventing the slippage with a strawberry (see Fig.~\ref{fig:setups}). The images are taken at ~30 cm distances from several viewpoints and at regular rotation intervals. We capture high-resolution images of $\SI{3264}{\px} \times \SI{2448}{\px}$ allowing for high-quality textures for realistic rendering. On average, we capture $\sim80$ images per strawberry. To match the appearance of the models to the field conditions, we capture the dataset outdoors using sunlight rather than artificial lights. We also use a white background to easily mask the fruit and improve the quality of the reconstructed shape. 

Our new dataset is significantly larger and offers a greater variability in the data distribution than previous work, such as~\cite{boli}, with increased texture and mesh quality. Of the 127 strawberries captured, 36 were from an unknown species bought at the farmer's market, and the rest were a mix between Zara and Katrina. Furthermore, we captured half of the strawberries at the unripe maturity stage and the other half fully ripe for greater variance in shape and appearance. The realism of the resulting rendered models is demonstrated in Fig.~\ref{fig:ours_examples}.

\begin{figure}[h]
    \centerline{
        \includegraphics[width=\columnwidth]{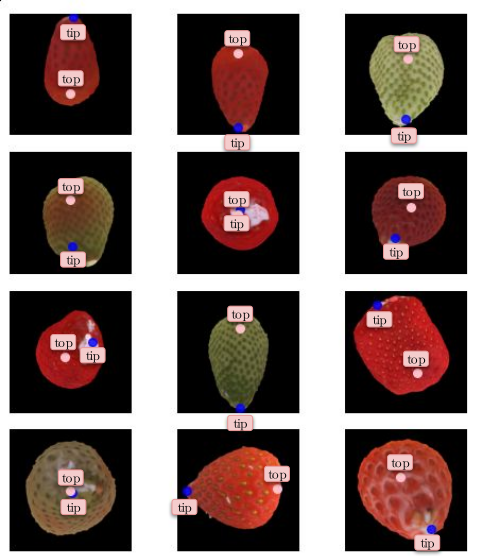}}
    \caption{Selected examples from the $Straw_{3D}$ dataset, consisting of post-processed images and annotations including top (pink) and tip (blue) key points.} \label{fig:set2annotationexample}
    \vspace{-0.2cm}
\end{figure}

To create the $Straw_{3D}$ dataset, each 3D berry is rendered in 84 different orientations resulting in 10668 individual views. The images were post-processed in a similar way as the $Straw_{2D}$ dataset resulting in squared and masked images with the resolution of $\SI{256}{\px} \times \SI{256}{\px}$. The annotation consists of the orientation ground truth (known due to the data being simulated) and the projected 3D key points (top and tip) on their image locations. The resulting image examples together with annotations are presented in Fig.~\ref{fig:set2annotationexample}.

Overall, we have 500 images representing 500 strawberry instances for $Straw_{2D}$ and 10668 image instances of 127 different strawberries for $Straw_{3D}$. Both datasets are annotated with ground truth key points, but only $Straw_{3D}$ has an associated orientation ground truth.

\subsection{Evaluation metrics and training details}
To evaluate the performance of our model for predicting the key point locations, we use the Euclidean distance $e=\sqrt{(x-x_{gt})^{2}+(y-y_{gt})^{2}}$ between the predicted $(x,y)$ and ground truth $(x_{gt},y_{gt})$ location expressed in pixels. For evaluating the predicted 3D pose for the strawberries from $Straw_{3D}$, we compute the angular distance between the predicted and ground truth pose. The direction vector $V = RV_{neutral}^\intercal$ where $V_{neutral}=\left[0,-1,0\right]$ and $R$ the rotation matrix obtained with $\phi$ and $\theta$. The angular error $\epsilon$ between the predicted $V_{pred}$ and ground truth $V_{gt}$ direction is then computed as $\epsilon = acos\left(V_{pred}V_{gt}\right)$. We use a recent method presented in \cite{wagner2021efficient} as the baseline for comparisons on the $Straw_{3D}$ dataset. We use a 10-fold cross-validation evaluation scheme for all datasets, with a dataset split of 90\% for training and 10\% for testing. For key points prediction, we train with a starting learning rate of \num{1e-4} for ${\sim}100$ epochs for $Straw_{2D}$ and ${\sim}6$ epochs for $Straw_{3D}$. We train our baseline~\cite{wagner2021efficient} and the $\theta$ regressor with a starting learning rate of \num{1e-5} until convergence (${\sim}6$ epochs).

\section{Results}
We first analyse the quality of key point detection, perform comparisons of the direct orientation computation and the supervised approach to the baseline method and present a qualitative analysis of the predictions for the estimated angles.

\begin{table}
    \begin{center}
        \begin{tabular}{|c|c|c|c|}
            \hline
            dataset  & $e_{tip}$ [px]  &  $e_{top}$ [px] & $e_{phi}$ \\
            \hline
            $Straw_{2D}$ & 9.25$\pm$4.20 & 14.91$\pm$ 9.45  &   $14.53^{\circ}\pm$ 39.85\\
            $Straw_{3D}$ & 7.61$\pm$2.61 & 8.60$\pm$3.21    &   $12.18^{\circ}\pm$ 27.05   \\
            
            \hline
        \end{tabular}
    \end{center}
    \caption{Median estimation errors for the top and tip key point locations in $Straw_{2D}$ and $Straw_3D$ datasets. As well as $\phi$ error, when computed from these key points.}\label{tab:kptserror}
     \vspace{-0.4cm}
\end{table}

Table~\ref{tab:kptserror} presents the median prediction errors of our key point detector for both datasets. For the in-field dataset $Straw_{2D}$, the results are around $\SI{9}{\px}$ (${\sim}4\%$ image size) for the tip and $\SI{19}{\px}$ (${\sim}7\%$ image size) for the top but with a high variance, which is still relatively good for imperfect fruit images collected from the field. This higher variance can be attributed to the quality of manual annotation, which was affected by the difficulties in identifying key points and strawberry shape for this dataset. Indeed with atypical shapes, and missing flesh $Straw_{2D}$, presents more challenges for annotation and prediction. With more accurate annotations present in the $Straw_{3D}$ dataset, however, the algorithm identifies the location of the key points more accurately which indicates that this is a critical consideration in training the key point detectors for real applications. Furthermore, the error in $\phi$ prediction ($e_{phi}$), shows a clear correlation between key-points accuracy and angle prediction, with the high-variance coming from difficult cases and ambiguously shaped strawberries. 

\begin{figure}[!ht]
 \vspace{-0.5cm}
    \centering
    \begin{subfigure}[t]{0.49\textwidth}
        \includegraphics[width=\textwidth]{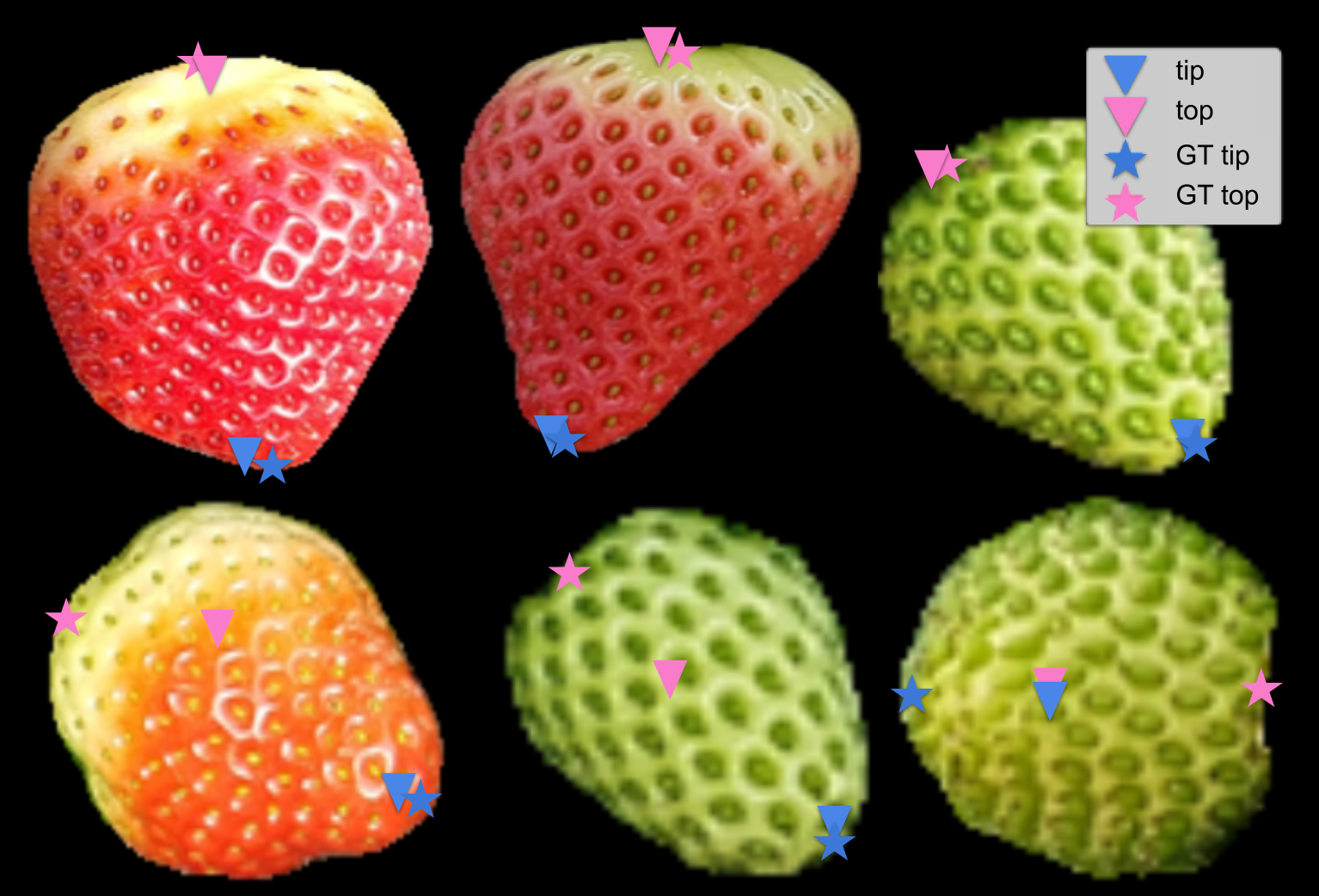} 
        \caption{}
        \label{fig:kptsset2}
    \end{subfigure}
    \hfill
     \begin{subfigure}[t]{0.49\textwidth}
        \includegraphics[trim={0 0 0 4em},clip,width=\textwidth]{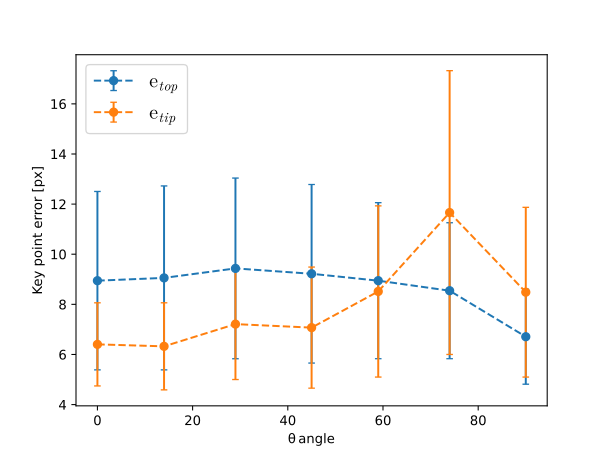} 
        \caption{}
        \label{fig:kperrorvsangle}
    \end{subfigure}
    \caption{(a) Examples of accurate (top row) and less accurate (bottom row) key point predictions from the $Straw_{2D}$ dataset.
    (b) Key point prediction error relative to the $\theta$ angle for $Straw_{3D}$. }
    \label{fig:my_label}
     \vspace{-0.5cm}
\end{figure}

We further show the correlation between key point localisation error and fruit orientation in Fig.~\ref{fig:kperrorvsangle} on $Straw_{3D}$. We can see that $e_{top}$ is consistent across different orientations but with a higher variance. $e_{tip}$  spikes mainly when the tip of the fruit points toward the camera and is harder to distinguish precisely (${\sim} 70^{\circ}$). Indeed, the top key point aligns with the centre of mass when $\theta$ gets close to $90^{\circ}$, while the tip key point is often displaced randomly due to the strawberry shape and growth. It is worth noting that more precise tip point estimation does not improve orientation results at lower values of $\theta$ as shown in Fig.~\ref{fig:distancestheta}.

Furthermore, we show in Fig.~\ref{fig:kptsset2} that the key points are predicted accurately in most of the cases. In $Straw_{2D}$, the errors are primarily due to inaccurate top prediction due to arguable and difficult to annotate precisely. We see in the last example also key points predicted on the centre of mass indicating a value of $\theta = 90^{\circ}$, probably due to the round shape and missing flesh information.
For $Straw_{3D}$, an higher imprecision with the tip prediction for high values of $\theta$ can be observed.

\begin{figure}[!ht]
    \begin{center}
        \begin{subfigure}{\linewidth}
            \includegraphics[width=\hsize]{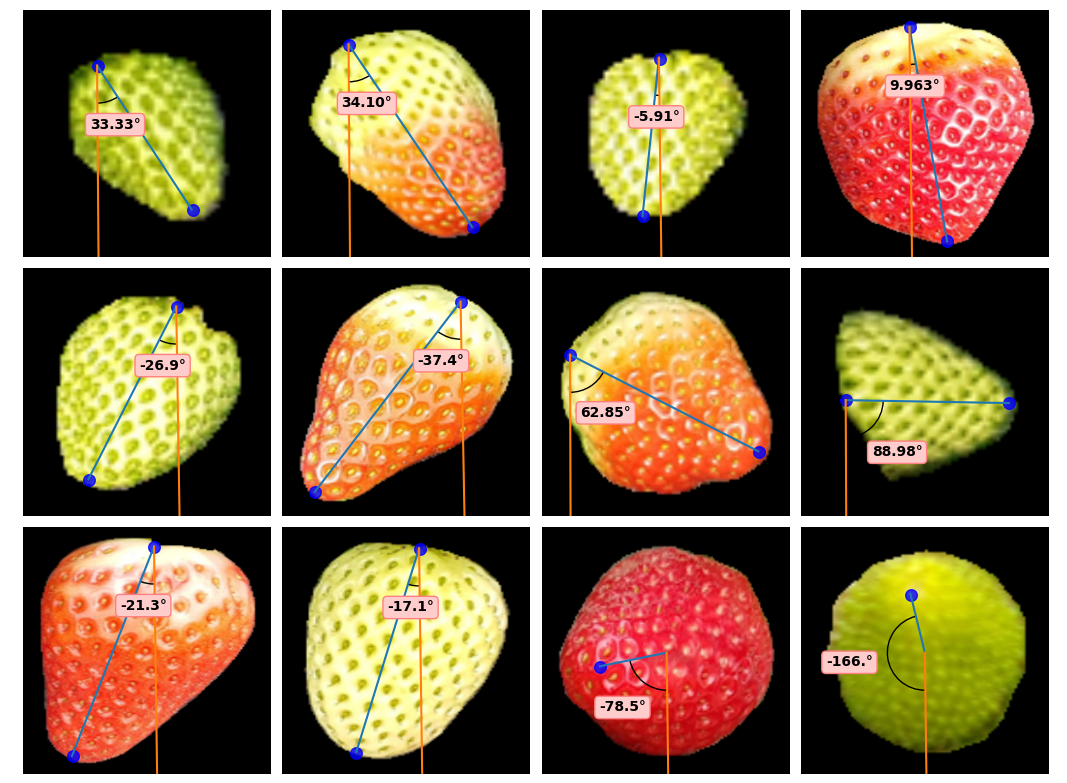}
            \caption{Computed $\phi$ values for selected strawberry instances.}
            \label{fig:pitch_exemp}
        \end{subfigure}
        \vfill
        \null
        \vfill
        \begin{subfigure}{\linewidth}
            \includegraphics[width=\hsize]{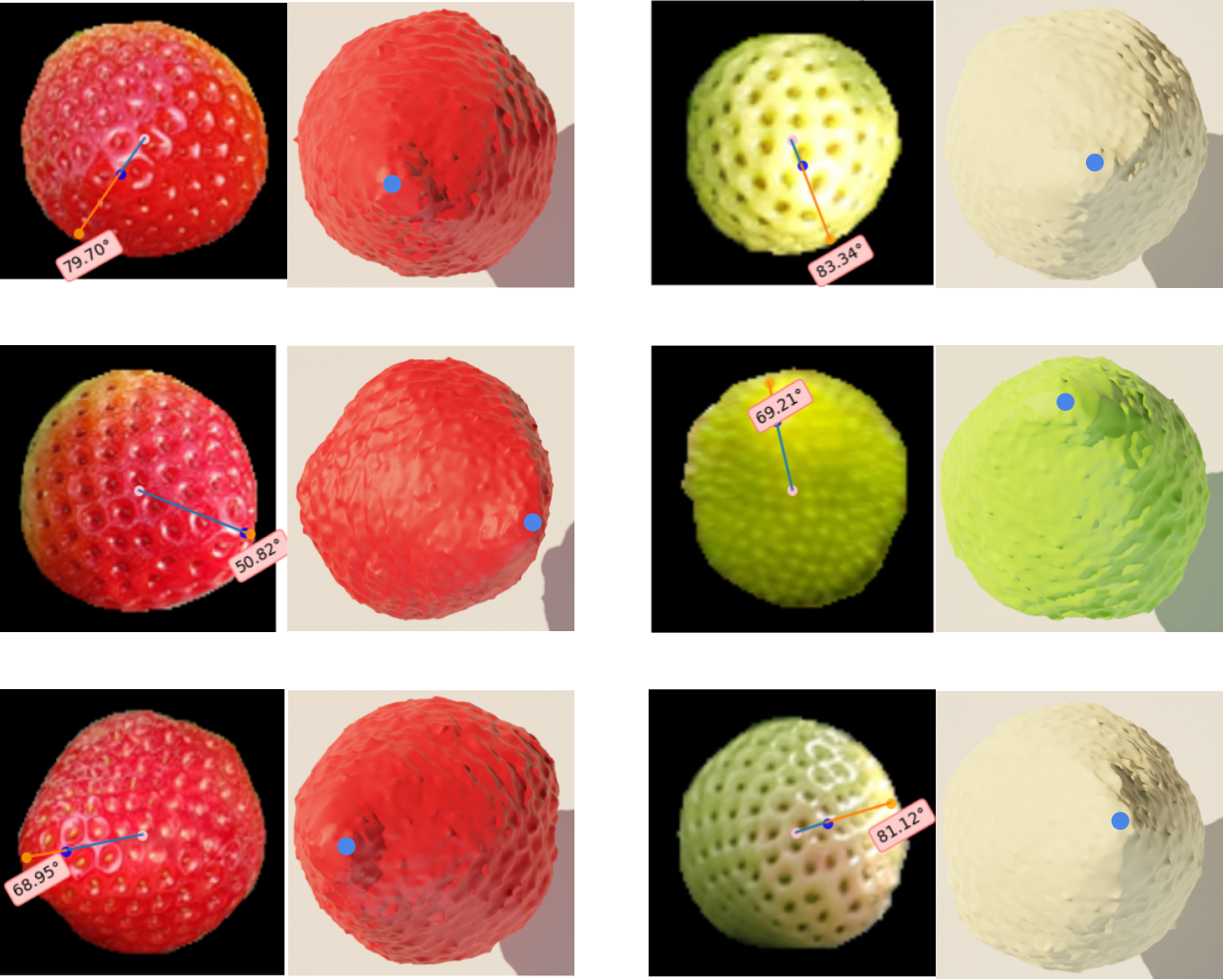}
            \caption{A comparison of estimation accuracy for $\theta$ and key point localisation between real strawberries (black background) and their 3D renders (white background). The tip key point is indicated in blue whilst the intersection with the contour is indicated in orange.}
            \label{fig:comparison_orientation}
        \end{subfigure}    
        \caption{Qualitative results obtained using the numerical method applied to the $Straw_{2D}$ dataset.}
        \label{fig:qualitative2d}
    \end{center}
    \vspace{-0.3cm}
\end{figure}

The example output from the key point detectors applied to the $Straw_{2D}$ dataset together with numerically calculated angles $\phi$ and $\theta$ is presented in Fig.~\ref{fig:pitch_exemp} and Fig.~\ref{fig:comparison_orientation} respectively. We indicate the tip key point on the rendered strawberries for $\theta$ to compare its localisation with respect to the reference strawberry. With accurate key points, in Fig.~\ref{fig:pitch_exemp} the numerical prediction of $\phi$ gives a clear indication of the fruit orientation within the image plane. The numerical computation of $\theta$ in Fig.~\ref{fig:comparison_orientation} shows that using the tip key point and contour of the fruit leads to accurate orientation estimates, with the template strawberry meshes precisely aligning with the target images.


   \begin{figure}[!t]
        \begin{center}
        \begin{subfigure}{\linewidth}
            \includegraphics[trim={0 0 0 4em},clip,width=\linewidth]{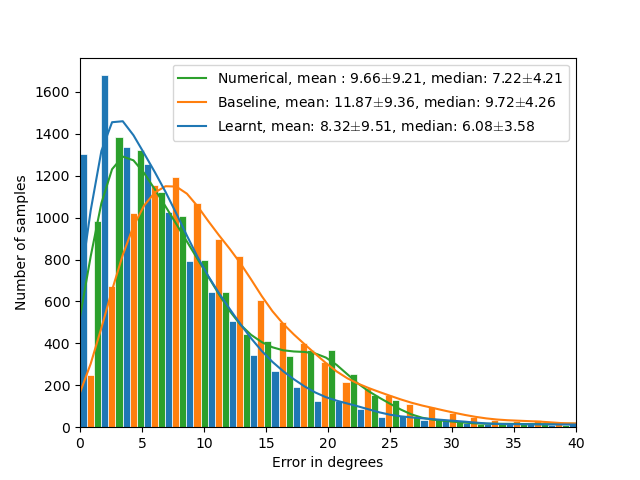}
            \caption{Distribution of angular errors and median and mean values.}
            \label{fig:angularerrorset1}
        \end{subfigure}
        \vfill
        \begin{subfigure}{\linewidth}
            \includegraphics[trim={0 0 0 3em},clip,width=\linewidth]{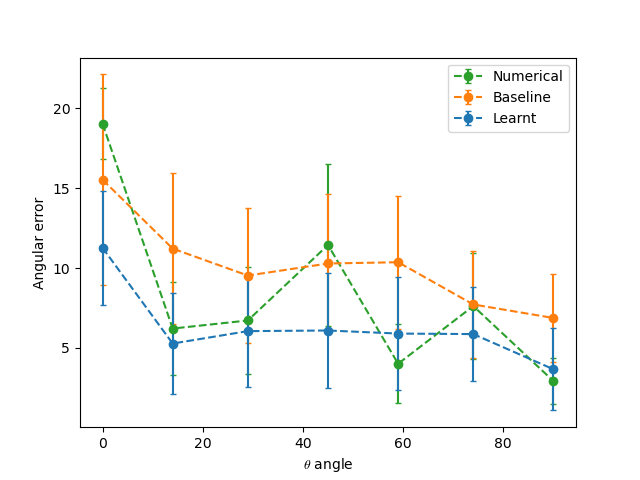}      
            \caption{Median estimation errors with respect to different values of $\theta$.}
            \label{fig:angularerrorset2}
        \end{subfigure}
        \caption{Quantitative evaluation of the proposed key point-based orientation estimation, supervised method and \cite{wagner2021efficient} on $Straw_{3D}$.}
        \label{fig:angularerrorset22}
        \end{center}
        \vspace{-0.2cm}
    \end{figure}

The comparison of the predicted poses from our proposed methods including direct numerical estimation and the learnt $\theta$ estimator to the baseline from~\cite{wagner2021efficient} is presented in Fig.~\ref{fig:angularerrorset22}. The evaluation was performed on the $Straw_{3D}$ dataset which includes the full pose annotation. The supervised orientation method performs significantly better with lower error ($\sim 8^{\circ}$), and less variance than the baseline ($\sim 11^{\circ}$). Furthermore, as displayed in Fig.~\ref{fig:angularerrorset2}, the numerical method shows viable results in general with worse performance for the most ambiguous poses $\theta=0^{\circ}$ and $\theta=49^{\circ}$ only. The latter is due to the ambiguous orientation lying in between the two parts stages of the formula used. This is a limitation of the numerical formula based on 2D images which is difficult to overcome without employing additional 3D shape information. The orientation prediction performs more accurately for all methods as $\theta$ gets closer to $90^{\circ}$. This is coherent with the lack of variation in appearance for values of $\theta$ bellow $30^{\circ}$ as presented in Fig.~\ref{fig:distancestheta}.

\begin{figure}[!ht]
    \includegraphics[width=\linewidth]{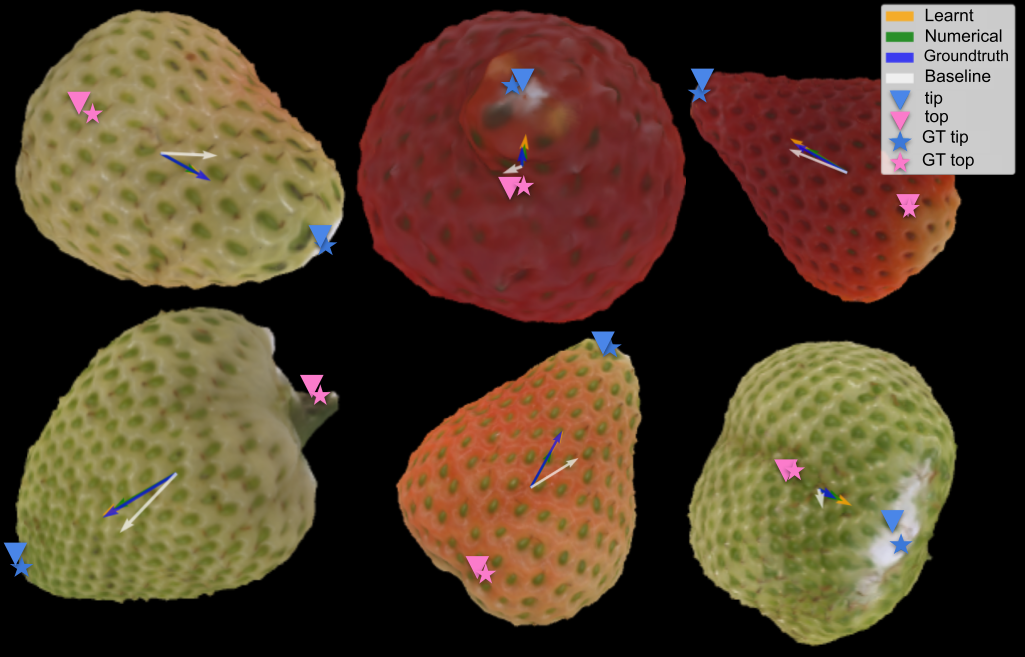} 
    \caption{Image projected orientation vector $V_{pred}$ (orange, green, white) for our method and baseline compared to the ground truth direction $V_{gt}$ (blue) on $Straw_{3D}$. The projected vector direction corresponds to $\phi$ and its length to $\theta$. The predicted (triangle) and ground truth (star) tip (blue) and top (pink) key points are also indicated.}
    \label{fig:orientationvisual}
    \vspace{-0.2cm}
\end{figure}

Finally, we visually compare the predicted orientation vectors for all three methods in Fig.~\ref{fig:orientationvisual}. The white direction vectors (\cite{wagner2021efficient}) confirm the results from Fig.~\ref{fig:angularerrorset1}, with the baseline always slightly off the ground truth and difficult to interpret in some cases. On the other hand, the key points give a better understanding when our method shows imprecision.


Our method is computationally very efficient and suitable for real-time robotic applications with inference times of $30.0$ ms for the key point detector and $1.4$ ms for the supervised regression of $\theta$. The performance is measured on an NVIDIA GeForce GTX 1880 Ti, with Intel(R) Core(TM) i7-7700K CPU and 16 GB memory.
    
\section{Conclusions and future work}
In this paper, we presented a novel approach for predicting the orientation of strawberries from single-view images. Our method exploits the key points and understanding of the fruit's shape, with two different techniques for predicting the rotation angles of the fruit. The experiments indicate that our approach achieves state-of-the-art results with average errors as low as $8^{\circ}$. In addition, our key point-based method leads to a better understanding of failure cases when compared to the baseline, clearly indicating sources of errors which are directly linked to mislocalised key points. Our method is suited for robotic strawberry harvesting where the fruit's orientation is important for effective end-effector motion planning.

Future work will include further optimisation of the key point prediction network and supervised regression of $\theta$. For example, sharing the weight from the feature extraction layers should reduce the training time and complexity of the models with the number of weights needed. The developed numerical $\theta$ computation was developed without considering the possible berry shapes which, if taken into account, should improve the results by improving the correlation between the key point location and orientation. Furthermore, this work can easily be extended to other crops by modifying the numerical formula. Future work would also include considering the external and self-occlusion of the key points, and adding uncertainty measures for hidden key points.


\bibliographystyle{IEEEtran}
\bibliography{IEEEexample}

\end{document}